\def\year{2017}\relax
\documentclass[letterpaper]{article}
\usepackage{aaai17}
\usepackage{times}
\usepackage{helvet}
\usepackage{courier}
\usepackage{multirow,varwidth}
\usepackage{enumerate}
\usepackage{enumitem}
\usepackage{color}
\usepackage{amssymb,amsmath,amsthm,enumitem}
\usepackage{graphicx}
\usepackage{lipsum}
\usepackage{hyperref}
\usepackage{textcomp}
\usepackage{tikz}
\usepackage{pgfplots}
\usepackage{caption} 
\begin{document}
\title{The Color of the Cat is Gray: \\1 Million Full-Sentences Visual Question Answering (FSVQA)}
\author{Andrew Shin, Yoshitaka Ushiku, Tatsuya Harada\\
The University of Tokyo\\
7 Chome-3-1 Hongo, Bunkyo\\
Tokyo 113-8654, Japan\\
}
\maketitle
\begin{abstract}
Visual Question Answering (VQA) task has showcased a new stage of interaction between language and vision, two of the most pivotal components of artificial intelligence. However, it has mostly focused on generating short and repetitive answers, mostly single words, which fall short of rich linguistic capabilities of humans. We introduce Full-Sentence Visual Question Answering (FSVQA) dataset (\textcolor{blue}{\url{www.mi.t.u-tokyo.ac.jp/static/projects/fsvqa}}), consisting of nearly 1 million pairs of questions and full-sentence answers for images, built by applying a number of rule-based natural language processing techniques to original VQA dataset and captions in the MS COCO dataset. This poses many additional complexities to conventional VQA task, and we provide a baseline for approaching and evaluating the task, on top of which we invite the research community to build further improvements.
\end{abstract}

\section{Introduction}

\begin{figure}[t]
\includegraphics[clip, trim=0cm 5.2cm 0cm 0cm, width=1.00\linewidth]{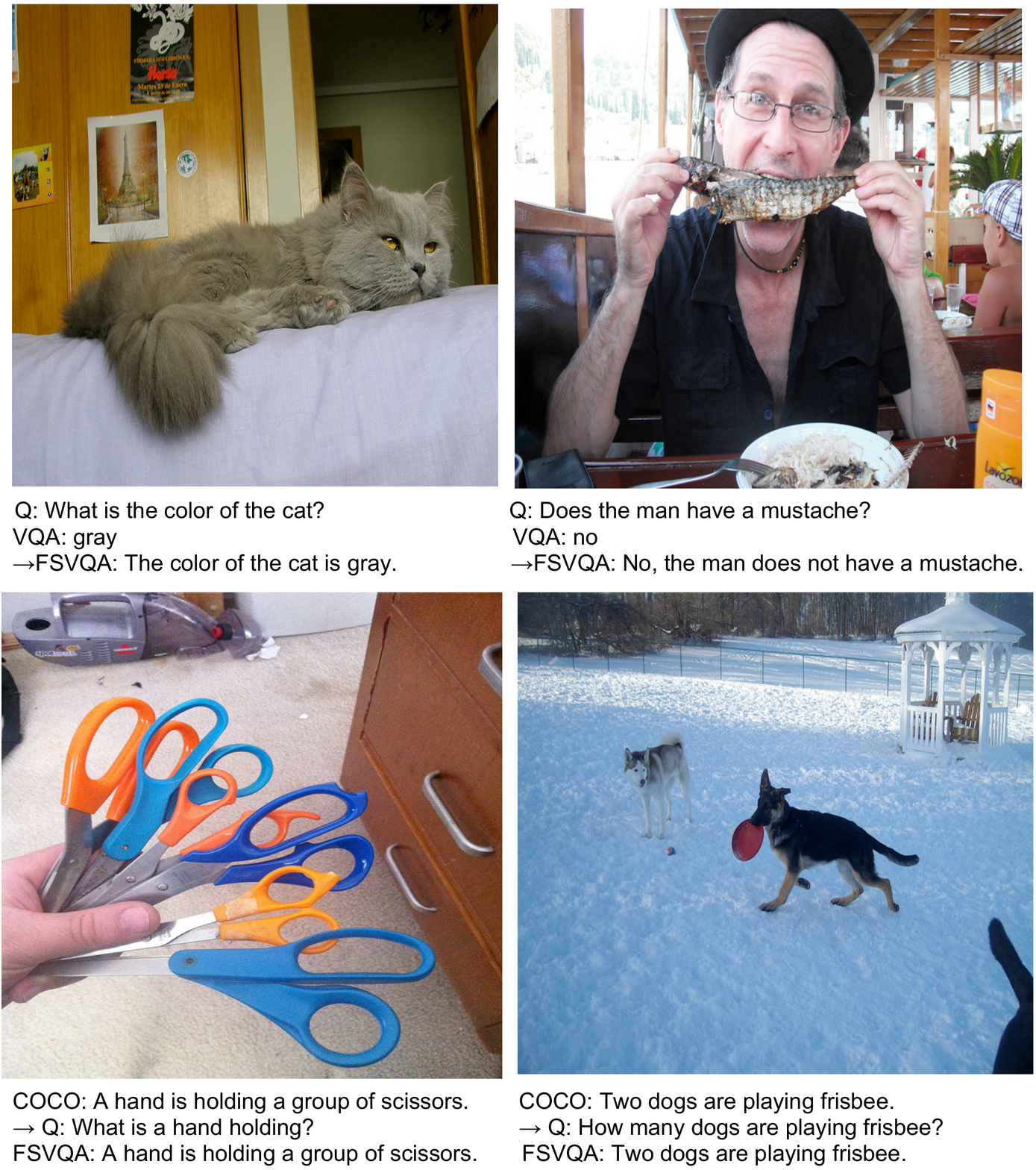}
\caption{Examples of FSVQA dataset. Top row shows examples of full-sentence answers generated by converting VQA questions and answers. Bottom row shows examples of questions and answers generated by converting MS COCO captions.}
\label{fig:teaser}
\vspace{-3ex}
\end{figure}

\noindent The research community in artificial intelligence (AI) has witnessed a series of dramatic advances in the AI tasks concerning language and vision in recent years, thanks to the successful applications of deep learning techniques, particularly convolutional neural networks (CNN) and recurrent neural networks (RNN). AI has moved on from naming the entities in the image \cite{Annot08,Annot09}, to describing the image with a natural sentence \cite{ShowAndTell,ShowAttendTell,Karpathy} and then to answering specific questions about the image with the advent of visual question answering (VQA) task \cite{VQA}. 

However, current VQA task is focused on generating a short answer, mostly single words, which does not fully take advantage of the wide range of expressibility inherent in human natural language. Just as we moved from merely naming entities in the image to description of the images with natural sentence, it naturally follows that VQA will also move towards full-sentence answers. One way to tackle this issue would be to apply appropriate linguistic rules \textit{after} a single-word answer is generated. However, previous works in natural language processing field have demonstrated that data-driven response generation achieves better performance than rule-based generation \cite{Ritter}. In other words, it is more efficient to provide data only once for pre-training than to parse and tag the text every time to apply universal rules. In addition, training with full-sentence answers provides an opportunity for the learning of complex morphological transformations along with visual and semantic understanding, which cannot be done with manual application of rules.

Learning and generating full-sentence answers will inevitably increase the number of distinct answers at an exponential scale; for example, thousands of samples with simple identical answer ``yes'' will be further divided into ``yes, the color of the car is red,''``yes, the boy is holding a bat,'' etc. Indeed, our FSVQA dataset contains almost 40 times more answers that are unique than the original VQA dataset. This poses additional challenges on top of original VQA task, since now it not only has to come up with the correct answer, but also has to form a full sentence considering how the words are conjugated, inflected, and ordered.

\setlength{\tabcolsep}{2pt}
\renewcommand{\arraystretch}{1.15}

 \begin{table*}[t]
\small
\begin{center}
\caption{General conversion rules for generating full-sentence answers.} 
\begin{tabular}{c|c|c|c|c}
\hline \bf Type & \bf Rule (Q$\rightarrow$A) & \bf Question & \bf Ans. & \bf Converted Ans. \\ \hline
\multirow{6}{*}{yes/no} & VB1+NP+VB2/JJ? & -- & -- & --\\ %\hline%\cline{2-8}% \cline{2-8}%complete
 & $\rightarrow$``\textit{Yes,}''+NP+\textit{conjug}(VB2/JJ,\textit{tense}(VB1))  \textbf{or},  &  \textit{Did he get hurt?} &\textit{yes} &\textit{Yes, he got hurt.}\\ 
& ``\textit{No,}''+NP+\textit{negate}(\textit{conjug}(VB2/JJ,\textit{tense}(VB1))) & \textit{Is she happy?} &\textit{no}&\textit{No, she is not happy.}\\ \cline{2-5}%\cline{2-8} 	%complete
& MD+ NP+VB?& -- & -- & --\\ %\hline%\cline{2-8}% \cline{2-8}%complete
 & $\rightarrow$``\textit{Yes,}''+NP+MD+VB  \textbf{or},  &\textit{Will the boy fall asleep?}  &\textit{yes} &\textit{Yes, the boy will fall asleep.}\\ 
&``\textit{No,}''+NP+\textit{negate}(MD)+VB & \textit{May he cross the road?} &\textit{no}&\textit{No, he may not cross the road.}\\ \hline%\cline{2-8} 	%complete
\multirow{4}{*}{number} &``\textit{How many}''+NP+\textit{/is/are}+EX?$\rightarrow$EX+\textit{is/are}+\textit{ans}+NP&\textit{How many pens are there?}&\textit{2}&\textit{There are 2 pens.}\\
&``\textit{How many}''+NP1(+MD)+VB(+NP2)? & -- & -- & --\\ 
& $\rightarrow$\textit{ans}(+MD)+VB(+NP2)& \textit{How many people are walking?}&\textit{3}&\textit{3 people are walking.}\\ %\cline{2-5}
&``\textit{How many}''+NP1+VB1/MD+NP2+VB2? & -- & -- & --\\ %\hline%\cline{2-8}% \cline{2-8}%complete
& $\rightarrow$NP2+(MD+VB2)/\textit{conjug}(VB2,\textit{tense}(VB1))+\textit{ans}+NP1& \textit{How many pens does he have?} &\textit{4}&\textit{He has 4 pens.}\\ \hline%\cline{2-8} 	%complete
\multirow{8}{*}{others} & WP/WRB/WDT+``\textit{is/are}''+NP?$\,\to\,$NP+``\textit{is/are}''+\textit{ans.}&\textit{Who are they?}&\textit{students}&\textit{They are students.}\\
&WP+NP+VP?$\,\to\,$\textit{ans.}+VP & \textit{What food is on the table?}& \textit{apple} & \textit{Apple is on the table.}\\ 
 & WDT+NP+VP(+NP2)?$\rightarrow$\textit{ans.}(+NP)+VP(+NP2) & \textit{Which hand is holding it?}&\textit{left} &\textit{Left hand is holding it.}\\ 
& WP/WDT+MD+VB?$\rightarrow$\textit{ans.}+MD+VB  & \textit{Who would like this?} &\textit{dog}&\textit{Dog would like this.}\\ %\cline{2-5}%\cline{2-8} 	%complete
& WP/WDT+MD+NP+VB?$\rightarrow$NP+MD+VB+\textit{ans.}& \textit{What would the man eat?} & \textit{apple} & \textit{The man would eat apple.}\\ 
 & WP/WDT+VP(+NP)?$\rightarrow$\textit{ans.}+VP(+NP)  &  \textit{Who threw the ball?} &\textit{pitcher} &\textit{Pitcher threw the ball.}\\ 
  & WP/WDT+VB1+NP+VB2? & -- &-- &--\\ 
& $\rightarrow$NP+\textit{conjug}(VB2,\textit{tense}(VB1))+\textit{ans.} &  \textit{What is the man eating?} &\textit{apple}&\textit{The man is eating apple.}\\ \hline%\cline{2-8} 	%complete
\end{tabular}
\label{table:scores}
\end{center}
\vspace{-2ex}
\end{table*}

\setlength{\tabcolsep}{3pt}

 \begin{table*}
\small
\begin{center}
\caption{General conversion rules for converting captions to questions.} 
\begin{tabular}{c|c|c|c}
\hline \bf Type & \bf Rule (C$\rightarrow$Q) & \bf Caption & \bf Question \\ \hline
\multirow{7}{*}{yes/no} & NP$\rightarrow$``\textit{Does it look like}''+NP& \textit{A man.} & \textit{Does it look like a man?} \\ %\hline%\cline{2-8}% \cline{2-8}%complete
& NP1 + VB (+NP2)& -- & -- \\ 
 & $\rightarrow$\textit{conjug}(``\textit{do}/\textit{be}'',\textit{tense}(VB))+NP1+\textit{conjug}(VB,present)(+NP2)?  &  \textit{A dog jumped.} &\textit{Did a dog jump?} \\ 
& NP1+MD+VB(+NP2)$\rightarrow$MD+NP1+VB(+NP2)?& \textit{A boy would hit the ball.} & \textit{Would a boy hit the ball?}\\ %\cline{2-5}%\cline{2-8} 	%complete
& EX+``\textit{is/are}''+NP$\rightarrow$``\textit{is/are}''+EX+NP?& \textit{There are cats.} & \textit{Are there cats?} \\ \cline{2-4} %\hline%\cline{2-8}% \cline{2-8}%complete
& same as above except \textit{replace}(NP1,random NP) \textbf{or},  & \textit{People are playing baseball.} &\textit{Are cats playing baseball?}  \\ 
& \textit{replace(VB,random VB)}& \textit{People are playing baseball.}  & \textit{Are people making coffee?} \\ \hline%\cline{2-8} 	%complete
\multirow{2}{*}{number} & EX+is/are+\textit{num}+NP$\rightarrow$How many+NP+is/are+EX? & \textit{There are two cats.} & \textit{How many cats are there?} \\ 
& \textit{num}+NP+VB$\rightarrow$\textit{replace}(\textit{num},how many) & \textit{Six cars are parked.} & \textit{How many cars are parked?} \\ \hline
\multirow{5}{*}{others} & NP1(+MD)+VB$\,\to\,$WP(+MD)+VB? \textbf{or},&\textit{A boy is running.}&\textit{Who is running?}\\
 & $\rightarrow$WP+MD/\textit{conjug}(``\textit{do}/\textit{be}'',\textit{tense}(VB))+NP+``\textit{do}/\textit{doing}'')?  &  \textit{A boy is running.} & \textit{What is a boy doing?} \\ 
&\textit{obj}/\textit{property}+VB(+NP)$\,\to\,$WP/WDT+\textit{category}+VP(+NP)&\textit{An apple is shown.}&\textit{What fruit is shown?} \\ 
& NP1+VB(+NP2)+IN+NP3  & -- &--\\ %\cline{2-5}%\cline{2-8} 	%complete
& $\rightarrow$WRB+\textit{conjug}(``\textit{do}/\textit{be}'',\textit{tense}(VB))+NP1+\textit{conjug}(VB,\textit{tense}(VB))?&\textit{Dogs run in a park.} & \textit{Where do dogs run?}\\  \hline%\cline{2-8} 	%complete
\end{tabular}
\label{table:scores}
\end{center}
\vspace{-3ex}
\end{table*}

We introduce Full-Sentence Visual Question Answering (FSVQA) dataset, built by applying linguistic rules to original VQA dataset at zero financial cost. We also provide an augmented version of FSVQA by converting image captions to question and answers. We examine baseline approaches, and utilize complementary metrics for evaluation, providing a guideline upon which we invite the research community to build further improvements.

Our primary contributions can be summarized as following: 1) introducing a novel task of full-sentence visual question answering, 2) building a large, publicly available dataset consisting of up to 1 million full-sentence Q\&A pairs, and 3) examining baseline approaches along with a novel combination of evaluation metrics.

\section{Related Work}
A number of datasets on visual question answering have been introduced in recent years \cite{DAQUAR,COCOQA}, among which \cite{VQA} in particular has gained the most attention and helped popularize the task. However, these datasets mostly consist of a small set of answers covering most of the questions, and most of the answers being single word. Our FSVQA dataset, derived from \cite{VQA}, minimizes such limitation by converting the answers to full-sentences, thus widely expanding the set of answers.

\cite{Sony} proposed multimodal compact bilinear pooling (MCB) to combine multimodal features of visual and text representations. This approach won the 1st place in 2016 VQA Challenge in real images category. \cite{Saito} proposed DualNet, in which both addition and multiplication of the input features are performed, in order to fully take advantage of the discriminative features in the data. This method won the 1st place in 2016 VQA Challenge in abstract scenes category.

\cite{Yang2016} was one of the first to propose attention model for VQA. They proposed stacked attention networks (SANs) that utilize question representations to search for most relevant regions in the image.  \cite{Noh} also built an attention-based model, which optimizes the network by minimizing the joint loss from all answering units. They further-proposed an early stopping strategy, in which overfitting units are disregarded in training. 

\cite{Lu} argued that not only visual attention is important, but also question attention is important. Co-attention model was thus proposed to jointly decide where to attend visually and linguistically. \cite{SNU} introduced multimodal residual network (MRN), which uses element-wise multiplication for joint residual learning of attention models. 

Most of the works above limited the number of possible answers, which was possible due to a small number of answers covering the majority of the dataset. Our FSVQA dataset imposes additional complexity to existing approaches by having a much larger set of possible answers, in which no small set of labels can cover the majority of the dataset.

\section{Dataset}
Collecting full-sentence annotations from crowd-sourcing tools can be highly costly. We circumvent this financial cost by converting the answers in the original VQA dataset \cite{VQA} to full-sentence answers by applying a number of linguistic rules using natural language processing techniques. Furthermore, we also provide an augmented version of dataset by converting the human-written captions provided in the MS COCO \cite{COCO}. We generated questions with a set of rules, for which the caption itself becomes the answer. Both versions of FSVQA dataset along with the features used in our experiment, as will be described in the Experiment Section, are publicly available for download. Note that, for both versions, only train and validation splits are provided, since test splits are not publicly available. Also, we only provide open-ended version, and do not provide multiple choice version.

\setlength{\tabcolsep}{9pt}

 \begin{table*}[t]
\small
\begin{center}
\caption{Statistics for VQA and two versions of FSVQA dataset. Note that statistics for VQA dataset were computed from the most frequent answers for each question.} 
\begin{tabular}{c|c|c|c|c|c|c}
\hline \multirow{2}{*}{\bf Dataset} &\multirow{2}{*}{\bf split}& \bf number of & \bf avg len& \bf number of & \bf number of & \bf top 1k  answers\\ 
& & \bf Q\&A pairs & \bf answer & \bf unique answers & \bf unique words & \bf coverage(\%) \\  \hline
\multirow{2}{*}{VQA} &train& 248,349 & 1.11  & 17,089 & 11,673 &86.74  \\ 
&val& 121,512  & 1.12 & 10,922 &8,078 & 86.56  \\ \cline{2-7}
\cite{VQA} &all& 369,861 & 1.11 & 22,497 & 14,528 & 86.56  \\  \hline
\multirow{3}{*}{FSVQA}&train & 248,349  & 6.06  & 173,016 &32,790 &13.02\\ 
&val& 121,512  & 6.04  & 92,135 & 22,199 & 13.27 \\ \cline{2-7}
&all& 369,861 & 6.05  & 248,223 &39,663 & 12.66  \\  \hline
\multirow{3}{*}{FSVQA aug.} &train& 662,462 & 10.38  & 579,761 & 57,468 & 4.98 \\ 
&val& 324,166  & 10.22  & 292,525 & 39,344 & 5.05  \\ \cline{2-7}
&all& 986,628  & 10.33  & 853,001 & 69,486 & 4.83  \\  \hline
\end{tabular}
\label{table:scores}
\end{center}
\vspace{-2ex}
\end{table*}

\subsection{Converting VQA}
VQA dataset comes with 10 annotations per question, and we chose one annotation per question that has the highest frequency as the single answer for corresponding question. If par, one annotation was randomly selected. 

VQA dataset mainly consists of three categories of questions; yes/no, number, and others. Table 1 summarizes the general conversion rule for generating full-sentence answers for each category, along with examples. Part-of-speech tag notation follows that of PennTree I Tags \cite{Penn}, except NP and VP refer to parse tree instead of than part-of-speech. \textit{tense}:$\mathbb{V}\,\to\,\mathbb{T}$ returns the tense of the input verb, \textit{conjug}:$\mathbb{V}\times\mathbb{T}\to\,\mathbb{V}$ conjugates the input verb to the input tense, where $\mathbb{V}$ is a space of verbs of all forms, and $\mathbb{T}$ is a set of tenses such that $\mathbb{T}$=\{\textit{past}, \textit{present}, \textit{future}, \textit{past perfect}, \textit{present perfect}, \textit{future perfect}\}, except it returns the input as is if the input is of JJ tag. \textit{negate}:$\mathbb{V}\,\to\,\mathbb{V}$ negates the input verb of given tense, and \textit{replace}(A,B) substitutes A by B. Parentheses indicate an optional addition, while A/B indicates an insertion of one of the two sides depending on the question. To briefly illustrate the process, these general conversion rules substitute the question phrase with the answer, and reorder the sentence with appropriate conjugation and negation.

While these general rules cover the majority of the questions, some types of questions require additional processing. For example, conditional statements with ``if,'' or selective statements such as ``who is in the picture, \textit{A} or \textit{B}?,'' can be handled by disregarding the sub-clauses and applying the rules to the main clause. Also, since VQA dataset consists of human-written natural language, it inevitably contains variations encompassing colloquialism, typos, grammar violations, and abridgements, which make it difficult to apply any type of general conversion rule. We either manually modify them, or leave them as they are.

\subsection{Converting Captions}
We also provide an augmented version of the dataset by converting the human-written captions for images into questions and answers. Apart from yes/no questions, the answers to the generated questions are the captions themselves, eliminating the burden for generating reliable answers. Most images in MS COCO come with 5 captions, and we generated distinct question for each caption, whose conversion rule is shown in Table 2.

We assigned at least two yes/no questions with one ``yes'' and one ``no'' to all images, roughly balancing the number of answers with ``yes,'' and ``no.'' Questions with affirmative answers involving ``yes'' were generated by simply rephrasing the caption such that the question asks to confirm the contents in the caption, for which the answer is an affirmative statement of the question (which is the caption itself), accompanied by ``Yes,'' in the beginning. 

Questions with non-affirmative answers involving ``no'' were generated by substituting parts of captions with random actions or agents. For agents, we randomly choose one class from 1,000 object classes of ILSVRC 2014 object detection task, and substitute it for given agent in the caption. For actions, we randomly choose one class from 101 action classes of UCF-101 \cite{UCF101} plus 20 classes of activities of daily living (ADL) from \cite{Ohnishi}, and substitute it for given verb phrase. Resulting questions are frequently of interesting non-sensical type, such as ``are the birds doing push-ups on the tree?," for which the answer is simply a negation of the question, accompanied by ``No,'' in the beginning. We expect that such set of questions and answers can also potentially help the learning of distinction between common sense and nonsense.

We also generated questions that are category-specific. When an object or a property of a specific category is referred to, we replace it with the category name, preceded by \textit{wh}-determiner or \textit{wh}-pronoun, to ask which object or property it belongs to. Our manually pre-defined categories included color, animal, room, food, transportation, and sport. Other rules are mostly reversed process of the conversion rules in Table 1, in which we deliberately mask certain content of the caption, replace it with appropriate question forms, and reorder the sentence.

\setlength{\tabcolsep}{9pt}

 \begin{table}
\small
\begin{center}
\caption{Number of unique answers for each category.} 
\begin{tabular}{c|c|c|c}
\hline \bf Dataset & \bf yes/no& \bf number & \bf others\\ \hline
VQA & 2 & 817 & 21,678  \\ 
FSVQA & 98,700 &27,119& 122,324\\ 
FSVQA aug. & 485,428 & 72,970 & 294,603 \\ \hline
\end{tabular}
\label{table:scores}
\end{center}
\vspace{-3ex}
\end{table}

\setlength{\tabcolsep}{7pt}

 \begin{table*}[t]
\small
\begin{center}
\caption{Performances of the generated answers on evaluation metrics} 
\begin{tabular}{c|c|c|c|cccc|c|c}
\hline \bf Dataset & \bf Model & \bf VQA Acc. & \bf FSVQA Acc. & \bf BLEU-1 & \bf BLEU-2 & \bf BLEU-3 & \bf BLEU-4 & \bf METEOR & \bf Cider  \\ \hline
\multirow{3}{*}{FSVQA} & LSTM Q+I & 32.54 & 16.74 & 56.9 & 41.5 & 31.8 & 23.9 & 0.233 &2.904 \\
& LSTM Q & 31.00 & 16.18 & 54.2 & 39.3 & 30.4 & 23.3 & 0.219  &2.755 \\ %complete
& Image & 9.72 &1.35& 15.4 & 2.6 & 0.8 & 0.2 & 0.101 & 0.412  \\ \hline%complete
\multirow{3}{*}{FSVQA aug.} & LSTM Q+I & 32.57 & 7.09 & 42.2 & 23.7 &15.1 & 10.2 & 0.150 &1.798\\ \ 
& LSTM Q & 31.53 & 6.69 & 40.5 & 22.9 & 14.6 & 10.0 & 0.141 &1.780 \\ %\cline{2-8} %complete
& Image  & 14.41 & 0.87 & 26.3 & 8.7 & 4.6 &  3.2 & 0.094 &0.601 \\	\hline  %complete
\end{tabular}
\label{table:scores}
\end{center}
\vspace{-2ex}
\end{table*}

\subsection{Statistics}
Table 3 shows statistics for original VQA dataset and two versions of FSVQA dataset. Both versions of FSVQA contain much longer answers on average, and the number of unique answers is more than ten times larger in the regular version and about 38 times larger in the augmented version compared to VQA dataset. Augmented version is longer since captions in MS COCO tend to be longer than questions in VQA. The size of vocabularies is also many times larger in both versions. Note that, consistently with the conversion process, only the most frequent answer from 10 answers per question was taken into consideration for VQA dataset's statistics. 

Comparing the coverage of 1,000 most frequent answers in the dataset shows even more striking contrast. In VQA dataset, 1,000 most frequent answers covered about 86.5\% of the entire dataset, so that learning a small set of frequent answers could perform reasonably well. In FSVQA, the coverage by 1,000 most frequent answers is merely 12.7\% and 4.8\% for regular and augmented version respectively, which is clearly much less than VQA dataset. Thus, it adds a critical amount of computational complexity, in which less frequent answers cannot easily be disregarded.

Table 4 shows the number of unique answers for each category. While FSVQA has much more answers in all categories, most striking example is in the yes/no category. VQA dataset essentially contains only two answers ``yes'' and ``no'' for yes/no questions (infrequent answers such as ``not sure'' were filtered out in the conversion process). In fact, answering with ``yes'' alone for all questions achieved 70.97\% accuracy for yes/no category in the original VQA dataset. On the contrary, FSVQA datasets contain approximately 49,000 times more answer and 240,000 times more answers for yes/no category in each version. It becomes clear again that FSVQA cannot be taken advantage of by manipulating a small set of frequent answers.

Figure 2 shows the percentage distribution of answers in the respective datasets for number of words in the answers. While over 90\% of the answers in VQA dataset are single words, both versions of FSVQA dataset show much smoother distribution over a wide range of number of words.

\section{Experiment}
\subsection{Setting}
We used 4096-dimensional features from the second fully-connected layer (fc7) of VGG \cite{vgg} with 19 layers, trained on ImageNet \cite{ImageNet}, as our image features. Words in the question were input to LSTM \cite{LSTM} one at a time as one-hot vector, where the dictionary contains only the words appearing more than once. Image features and question features are then mapped to common embedding space as a 1,024-dimensional vector. Batch size was 500 and training was performed for 300 epochs.

We trained only with the answers that appear twice or more in train split, as using all unique answers in the dataset fails to run, with required memory far beyond the capacity of most of the contemporary GPUs, NVIDIA Tesla 40m in our case. 20,130 answers appear more than once in regular version, covering 95,340 questions from 62,292 images, and 23,400 answers appear more than once in the augmented version, which cover 105,563 questions from 64,060 images. This is only about 25\% and 15.5\% of the entire train split in respective version, which again shows a striking contrast with the original VQA dataset, in which only 1,000 answers covered up to 86.5\% of the dataset.

Following \cite{VQA}, we examined the effect of removing images, since questions alone may frequently provide sufficient amount of clue to correct answers. Training procedure is identical as above, except only question features are mapped to the common embedding space since there are no image features.

Conversely, we also examined an approach where only image features are concerned. This requires a slightly different training procedure, as it does not involve a series of one-hot vector inputs. We followed the conventional approach used in image captioning task \cite{ShowAndTell}, where the image features are fixed, and a stack of LSTM units learns to generate the ground truth captions. Each LSTM unit generates one word at a time, which in turn enters the next LSTM unit. The only difference in our case is that the ground truth captions are replaced by full-sentence answers.

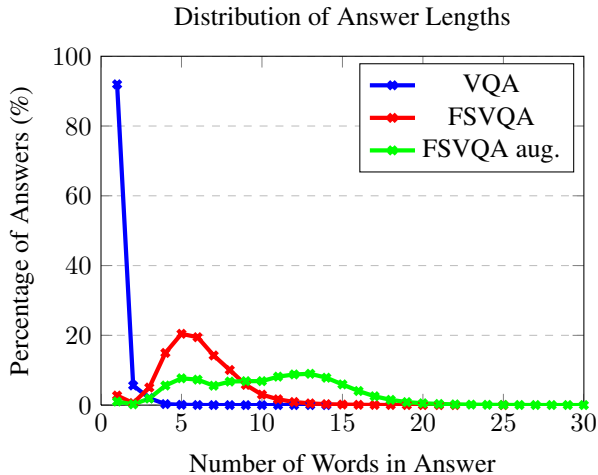
\begin{figure}
\begin{tikzpicture}
\centering
\pgfplotsset{
}
%\begin{subfigure}
\begin{axis}[
      width=0.45\textwidth,
      height=0.35\textwidth,
title=Distribution of Answer Lengths,
ylabel near ticks, yticklabel pos=left,
  ymin=0, ymax=100,
  xmin=0,xmax=30,
  xlabel=Number of Words in Answer,
  ylabel=Percentage of Answers (\%),
    ymajorgrids=true,
    grid style=dashed,
]
\addlegendentry{VQA}
\addplot[ultra thick,mark=x,blue]
  coordinates{
(1,91.95)
(2,5.63)
(3,1.99)
(4,0.29)
(5,0.09)
(6,0.03)
(7,0.01)
(8,0)
(9,0)
(10,0)
(11,0)
(12,0)
(13,0)
(14,0)
};% \label{plot_three}
\addplot[ultra thick,mark=x,red]
  coordinates{
(1,2.73)
(2,0.59)
(3,5.04)
(4,14.96)
(5,20.44)
(6,19.46)
(7,14.23)
(8,10.07)
(9,5.80)
(10,3.00)
(11,1.66)
(12,0.89)
(13,0.49)
(14,0.27)
(15,0.16)
(16,0.09)
(17,0.05)
(18,0.03)
(19,0.01)
(20,0)
(21,0)
(22,0)
(25,0)
}; %\label{plot_three}
\addlegendentry{FSVQA}
\addplot[ultra thick,mark=x,green]
  coordinates{
(1,1.02)
(2,0.22)
(3,1.89)
(4,5.61)
(5,7.68)
(6,7.30)
(7,5.56)
(8,6.75)
(9,6.87)
(10,6.87)
(11,8.16)
(12,8.79)
(13,8.98)
(14,7.85)
(15,5.93)
(16,4.04)
(17,2.53)
(18,1.51)
(19,0.86)
(20,0.53)
(21,0.33)
(22,0.21)
(23,0.14)
(24,0.09)
(25,0.06)
(26,0.04)
(27,0.03)
(28,0.02)
(29,0.02)
(30,0.01)
}; %\label{plot_three}
\addlegendentry{FSVQA aug.}
\end{axis}
%\end{subfigure}
\end{tikzpicture}
%\caption{Performances of Combination of Features}
%\end{subfigure}
\caption{Distribution of answer lengths as the number of words over each dataset.}
\vspace{-3mm}
\end{figure}

\begin{figure*}[t]
\begin{center}
\includegraphics[clip, trim=0cm 0.2cm 0cm 0cm, width=0.95\linewidth]{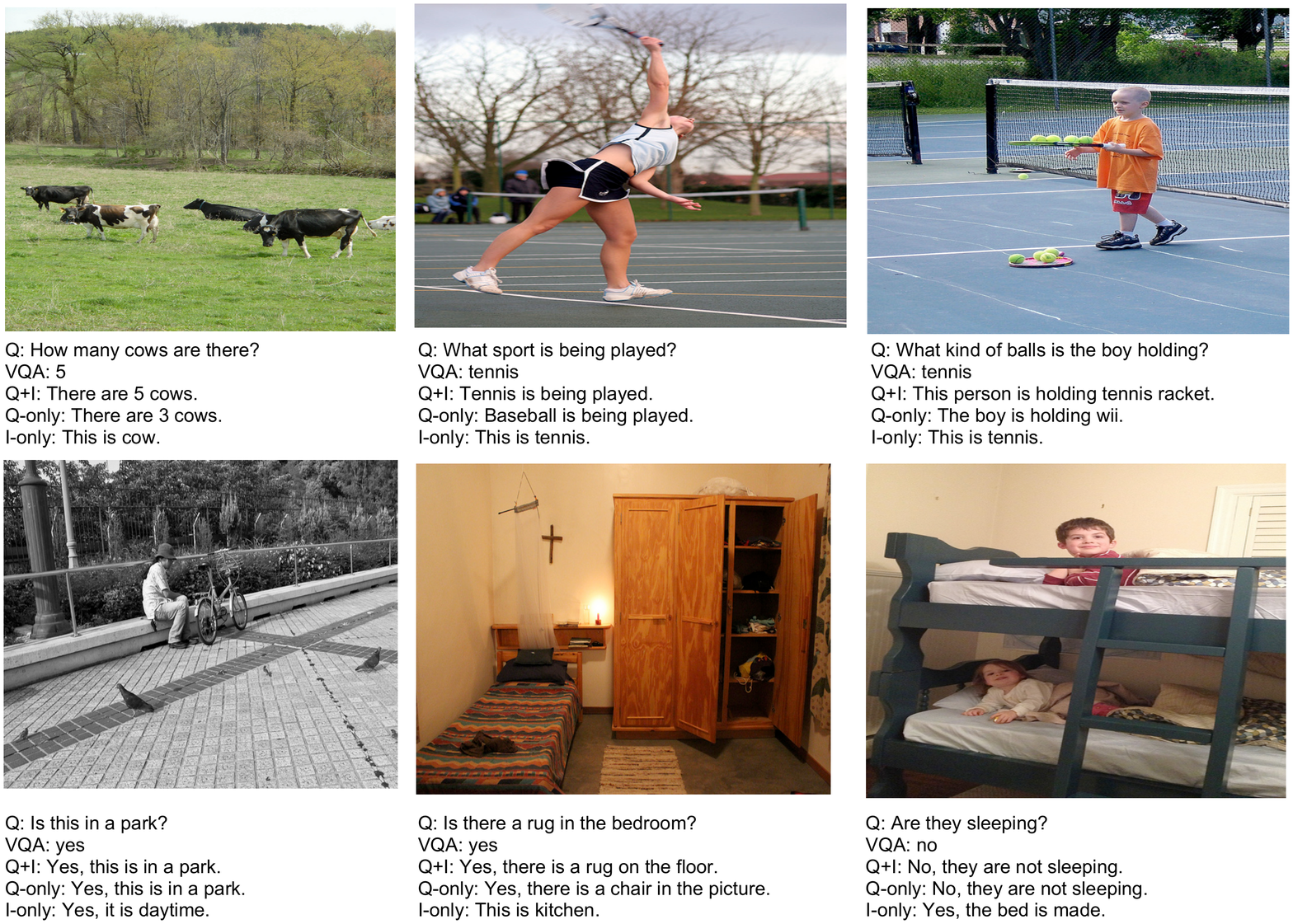}
\caption{Examples of questions and generated answers by each baseline approach, along with ground truth answer from the original VQA dataset.}
\label{fig:example}
\end{center}
\vspace{-2ex}
\end{figure*}

\subsection{Evaluation}
\subsubsection{Metrics}
Evaluating the results from the experiments also poses a challenge. Since original VQA dataset consisted mostly of short answers, evaluation was as simple as matching the results with ground truths, and yielding the percentage. Yet, since we now have full-sentence answers, simply matching the results with ground truths will not be compatible. For example, ``yes, the color of the cat is red'' for the question in which the original ground truth is ``yes,'' will be classified as incorrect using the current evaluation tool for original VQA. 

We thus come up with a set of mutually complementary ways of evaluating full-sentence answers. First, we employ the frequently used evaluation metrics for image captioning task, namely BLEU \cite{BLEU}, METEOR \cite{Meteor}, and CIDEr \cite{Cider}. The goal is to quantify the overall resemblance of the results to the ground truth answers. 

However, higher performance on these evaluation metrics does not necessarily imply that it is a more accurate answer. For instance, ``the color of the car is red'' will have higher scores than ``it is blue'' for a ground-truth answer ``the color of the car is blue,'' due to more tokens being identical. Yet, the former is clearly an incorrect answer, whereas the latter should be considered correct. In order to overcome this drawback, we also employ a simple complementary evaluation metric, whose procedure is as follows: we examine whether the short answer from the original dataset is present in the generated result, and extract the short answer if present. If not, we leave the answer blank. Extracted terms in this way are tested with the evaluation tool for original VQA. Using the previous example, the rationale is that as long as the original short answer ``blue'' is present in the generated result, it can be assumed that the answer is correct. We refer to this metric as VQA accuracy. Note that, for augmented version, generated answers for only the original subset are extracted to measure VQA accuracy, since there are no ground truth VQA answers for the augmented segment.

However, there also exist cases in which VQA accuracy can be misleading, since the rest of the context may not be compatible with the question. For example, ``yes, the color is blue.'' will be considered correct if the original answer is ``yes,'' but it should not be considered correct if the question was ``is it raining?'' In fact, this was one of the underlying concerns in the original VQA dataset, since we cannot be sure whether ``yes'' or ``no'' was generated in the right sense or purely by chance. In order to alleviate this issue, we also report FSVQA accuracy, which is the percentage in which the ground truth answer in FSVQA dataset contains the generated answer. Since the answers have to be matched at the sentence level, it assures us with high confidence that the answer was correct in the intended context.

\subsubsection{Results \& Discussion}
Results for all metrics are shown in Table 5. Note that evaluation is performed on the results for validation split, since ground truths for test split are not publicly available. While each metric is concerned with slightly different aspect of the answers, results shows that they generally tend to agree with each other. Figure 3 shows examples of generated full-sentence answers for each model, along with the ground truth answer from the original VQA dataset. 

As expected, answers generated from using both question and image features turn out to be most reliable. Answers from question features alone result in answers that match the questions but are frequently out of visual context given by the image. Likewise, answers generated from image features alone fit the images but are frequently out of textual context given by the question. It is notable that using image features alone performs very poorly, whereas using question features alone results in performances comparable to using both features. One plausible explanation is that, since using image features alone always generates the same answer for the same image regardless of the question, it can only get 1 out of k questions correctly at best, where k is the number of questions per image. On the contrary, using question features alone essentially reduces the problem to a semantic Q\&A task, which can be handled one at a time. This tendency is consistent with the results reported in \cite{VQA}. It must nevertheless be reminded that the best performances in both \cite{VQA} and our experiment were achieved with the presence of both visual and textual clues.

%Finally, note that VQA accuracy (32.54) for the best-performing model of LSTM Q+I is much lower than the baseline approach for original VQA dataset, in which 

\section{Conclusion}
We introduced FSVQA, a publicly available dataset consisting of nearly 1 million pairs of questions and full-sentence answers for images, built by applying linguistic rules to existing datasets. While pushing forward the VQA task to a more human-like stage, it poses many extra complexities. We examined baseline approaches for tackling this novel task. Applying some of the successful approaches from the original VQA task, such as attention mechanism, will be an intriguing and important future work. Whether generative approach can play more role in the future, as the number of answers grows larger and classification approach becomes less efficient, is also of interest. We invite the research community to come up with an innovative and efficient way to improve the performance on FSVQA.
 
\section*{Acknowledgement}
This work was funded by ImPACT Program of Council for Science, Technology and Innovation (Cabinet Office, Government of Japan).

\bibliographystyle{aaai}
\bibliography{bbb}

\end{document}